\documentclass{article}

\usepackage{arxiv}

\usepackage[utf8]{inputenc} 
\usepackage[T1]{fontenc}    
\usepackage{hyperref}       
\usepackage{url}            
\usepackage{booktabs}       
\usepackage{amsfonts}       
\usepackage{nicefrac}       
\usepackage{microtype}      
\usepackage{lipsum}
\usepackage{graphicx}
\graphicspath{ {./images/} }
\usepackage{subfigure}
\usepackage{algorithm}
\usepackage{algorithmic}
\usepackage{multirow}
\usepackage{float}
\usepackage{amsmath}
\usepackage{amssymb}

\title{Can LLM be a Good Path Planner based on Prompt Engineering? Mitigating the Hallucination for Path Planning}

\author{
Hourui Deng \\
  College of Computer Science\\
  Sichuan Normal University\\
  Chengdu, China \\
  \texttt{herry.liquor@gmail.com} \\
   \And
 Hongjie Zhang\thanks{Corresponding Author} \\
  College of Computer Science\\
  Sichuan Normal University\\
  Chengdu, China \\
  \texttt{zhanghongjie@sicnu.edu.cn} \\
  \And
 Jie Ou \\
  School of Information and Software Engineering\\
  University of Electronic Science and Technology of China\\
  Chengdu, China \\
  \texttt{oujieww6@gmail.com} \\
 \And
 Chaosheng Feng \\
  College of Computer Science\\
  Sichuan Normal University\\
  Chengdu, China \\
  \texttt{csfenggy@sicnu.edu.cn} \\
}

\begin{document}
\maketitle
\begin{abstract}
Spatial reasoning in Large Language Models (LLMs) is the foundation for embodied intelligence. However, even in simple maze environments, LLMs still encounter challenges in long-term path-planning, primarily influenced by their spatial hallucination and context inconsistency hallucination by long-term reasoning. To address this challenge, this study proposes an innovative model, Spatial-to-Relational Transformation and Curriculum Q-Learning (S2RCQL). To address the spatial hallucination of LLMs, we propose the Spatial-to-Relational approach, which transforms spatial prompts into entity relations and paths representing entity relation chains. This approach fully taps the potential of LLMs in terms of sequential thinking. As a result, we design a path-planning algorithm based on Q-learning to mitigate the context inconsistency hallucination, which enhances the reasoning ability of LLMs. Using the Q-value of state-action as auxiliary information for prompts, we correct the hallucinations of LLMs, thereby guiding LLMs to learn the optimal path. Finally, we propose a reverse curriculum learning technique based on LLMs to further mitigate the context inconsistency hallucination. LLMs can rapidly accumulate successful experiences by reducing task difficulty and leveraging them to tackle more complex tasks. We performed comprehensive experiments based on Baidu's self-developed LLM: ERNIE-Bot 4.0. The results showed that our S2RCQL achieved a 23\%--40\% improvement in both success and optimality rates compared with advanced prompt engineering.
\end{abstract}


\section{Introduction}
Large language models are remarkable artificial intelligence (AI) technology that has gained remarkable attention in various fields. The LLMs have implemented Artificial Intelligence-Generated Content (AIGC) through massive corpora and advanced transformer frameworks. With the support of various prompt engineering, they have demonstrated a considerable level of intelligence and accomplished a wide range of decision-making tasks, such as mathematical reasoning ~\cite{Imani}, embodied AI agent ~\cite{Dorbala}, UAV control ~\cite{Piggott}, and complete open-world game Minecraft ~\cite{Zhu} using chain-of-thought technology. However, LLMs exhibit significant limitations in spatial reasoning and long-term planning, which caused by their spatial hallucination and context inconsistency hallucination by long-term reasoning.

\begin{figure}[!ht]
	\centering
	\subfigure[This is a maze that contains three forbidden zones.]{\label{fig:example:a}
		\includegraphics[width=1.05in]{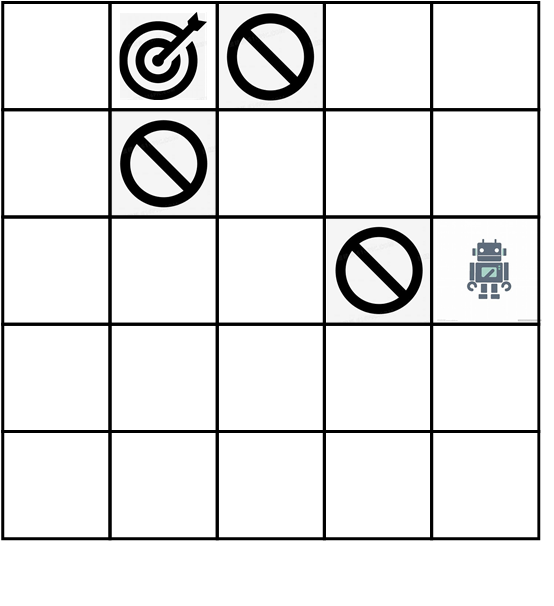}}
  \hspace{0.1mm}
	\subfigure[The heatmap of the path using CoT as a prompt to solve this maze.]{\label{fig:example:b}
		\includegraphics[width=1.7in]{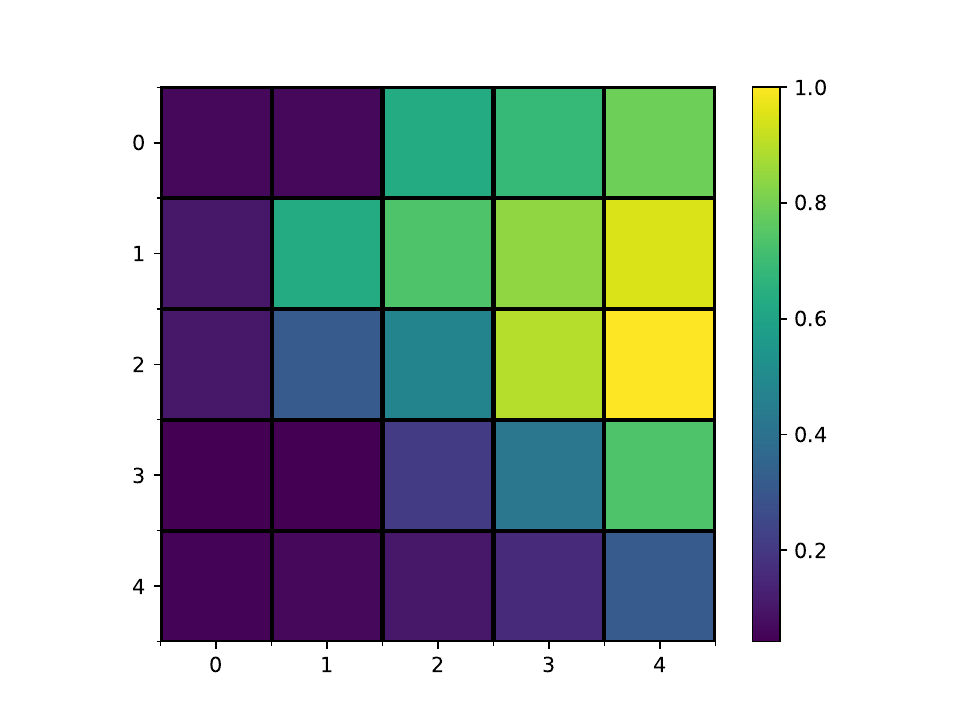}}
  \hspace{0.1mm}
	\subfigure[The heatmap of the path using Rememberer to solve this maze step by step.]{\label{fig:example:c}
		\includegraphics[width=1.7in]{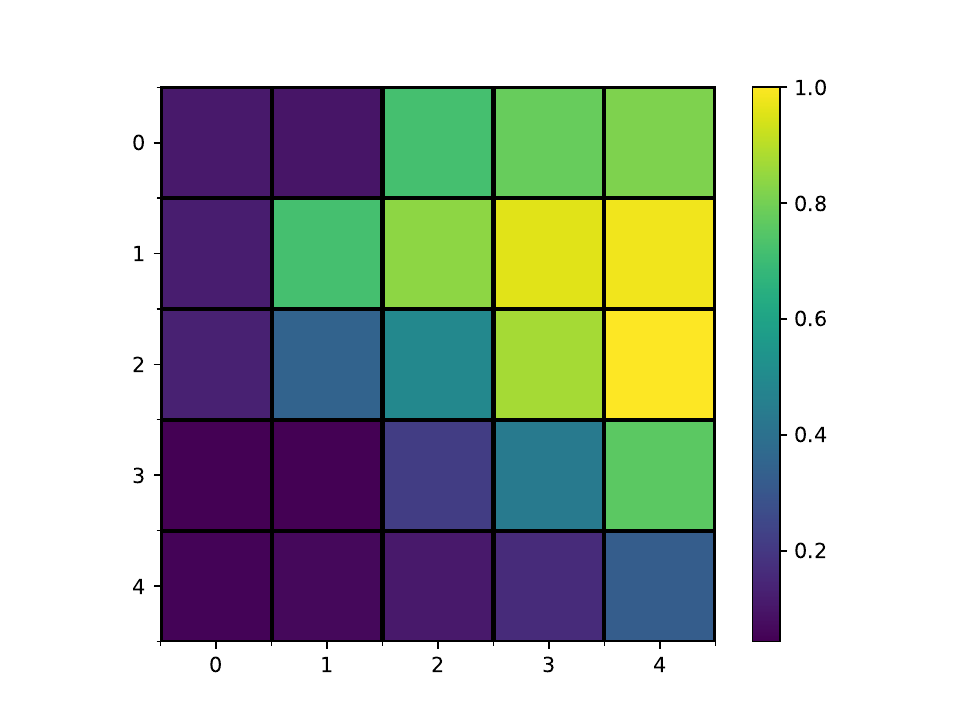}}
	\caption{An example of a maze. Solving this maze path planning task is challenging using both CoT and Rememberer, an LLMs with RL method.}
	\label{fig:example}
\end{figure}

Many studies have proposed various solutions to address hallucination problems, mainly focusing on three aspects: instruction fine-tuning, prompt engineering, and reinforcement learning. Instruction fine-tuning involves parameter adjustment of pre-trained LLMs, encompassing dataset curation and neural network training, to enhance performance on the specific task~\cite{Ye}. However, the significant computational costs required for fine-tuning LLMs pose a challenge for rapid expansion to new tasks. Prompt engineering aims to improve the inference accuracy of LLMs by designing instructions that guide the models to reason according to specific requirements. Advanced techniques in this domain include chain-of-thought (CoT)~\cite{Wei,Chu,Feng}, tree-of-thought (ToT)~\cite{Yao}, graph-of-thought (GoT)~\cite{Besta,Yao Y}, chain of experts (CoE)~\cite{Xiao Z}, ReAct~\cite{Yao S,Aghzal}, and Reflexion~\cite{Shinn}. Reinforcement learning (RL) has long been an effective technique for addressing complex planning problems by allowing an agent to interact with its environment through trial and error. This RL model continuously adjusts its strategies to achieve optimal path planning. By combining RL with LLMs, RL can reduce the cost of exploration~\cite{Zhang D,Carta T}. However, prompt and RL models perform poorly in spatial reasoning tasks, particularly maze path planning. As shown in Figure~\ref{fig:example:a}, this maze contains three forbidden zones. Therefore, a path must be planned from the starting point to the ending point. However, finding the shortest path at a glance is intuitive for humans. Still, when using CoT (Figure~\ref{fig:example:b}) and Rememberer~\cite{Zhang D}(Figure~\ref{fig:example:c}), agents often get stuck and return with a failure.

We have analyzed the motives behind LLMs' inadequate understanding of spatial relationships and their tendency to navigate naive paths in the direction of the shortest straight-line distance while ignoring obstacles. We propose an innovative method called Spatial-to-Relational Transformation and Curriculum Q-Learning (S2RCQL) to improve the LLM's performance and solve maze problems. We have introduced the Spatial-to-Relational transformation to address the issue of LLM's spatial hallucination. This transformation converts implicit spatial relationships into an explicit entity relation, describing the connectivity of paths through the relationships between nodes. Then, we introduced a Q-learning-assisted path-planning algorithm for LLMs to eliminate the LLMs' context inconsistency hallucination by long-term reasoning. We guided LLMs to avoid dead end by inserting Q-values into the prompts. Finally, we have designed a reverse curriculum learning algorithm to mitigate the context inconsistency hallucination further. This algorithm gradually increases task difficulty, allowing LLMs to reduce the number of reasoning steps. We performed extensive experiments based on Baidu's self-developed LLM: ERNIE-Bot 4.0. The results indicate that our \textbf{S2RCQL} achieves a significant improvement of 23\%--40\% in success and optimality rates compared to the state-of-the-art CoTs baselines. This study offers several contributions:
\begin{enumerate}
\item This study is the first to propose converting spatial path planning tasks into entity relations for path planning. As a result, we have designed the Spatial-to-Relational transformation, which has been successfully applied to maze navigation tasks.
\item We have also proposed a path-planning algorithm based on LLMs and reverse curriculum Q-learning, representing a step forward in addressing LLM's context inconsistency hallucination.
\item We performed comprehensive experiments based on the ERNIE-Bot model to verify the reliability and effectiveness of our algorithm.
\end{enumerate}

\section{Methodology}
\label{sec:headings}
\subsection{Overview}
\begin{figure}[!ht]
\includegraphics[width=\textwidth]{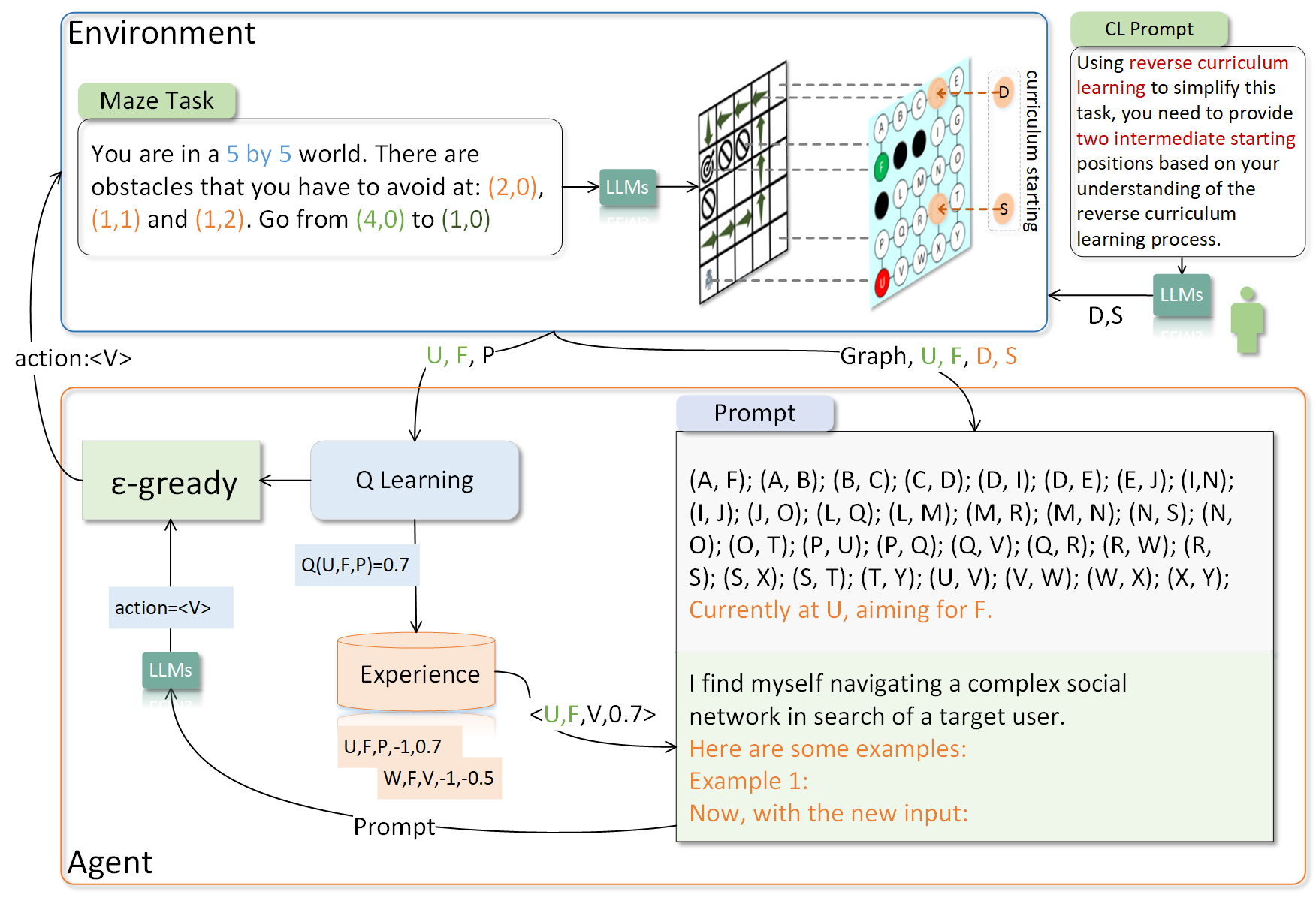}
\caption{This diagram provides an overview of our approach. First, we convert arbitrary text maze descriptions into entity relations using LLMs and Python code. Then, we combined the Q-learning and LLMs to select actions through $\epsilon-greedy$ with reverse curriculum learning.} \label{fig:overview}
\end{figure}

Figure~\ref{fig:overview} presents an overview of \textbf{S2RCQL}. Generally, \textbf{S2RCQL} comprises three main components: the environment, the agent, and the course module. The agent continuously interacts with the environment, seeking the shortest path from the starting point to the endpoint through trial and error. The environment module declares the maze in text with coordinates, including the maze size, obstacles, starting point, and goal. Then, we extract this format information using LLMs. To facilitate Python parsing, we control LLM output in \textit{JSON} format. Moreover, we automatically construct a graph from the \textit{JSON} response of LLMs to explicitly represent the maze connectivity, facilitating the LLMs' reasoning. In the agent module, we construct a state description of the current maze, including the graph structure and node. In addition, we retrieve the most similar experience and its corresponding Q-value from the experience replay buffer. \textbf{These elements are then used to create a few-shot example concatenated with the current maze} to form the final prompt. The agent outputs the final action, the next-hop node, based on the $\epsilon-greedy$ algorithm and updates the environmental state. This process is repeated iteratively. In the top right corner of Figure~\ref{fig:overview}, we identify the reverse course generation module. We construct intermediate starting points from the current graph based on hand-craft or LLMs. These starting points can reduce task difficulty and eliminate the context inconsistency hallucination by reducing the number of reasoning steps.

\subsection{Spatial-to-Relational Transformation}
Researchers from Google DeepMind and University College London have comprehensively analyzed LLMs' capabilities in performing potential multi-step reasoning~\cite{Yang S}. Multi-step reasoning requires models to retrieve relevant information sequentially and piece it together to solve problems or respond to queries. As a result, the relevant information required for LLMs' reasoning is crucial. CoT and ToT techniques involve generating intermediate reasoning processes using LLMs, including relevant information and piecing together this information to generate answers. In our maze planning task, we automatically convert map information into an entity relation format, enabling the relevant information required for LLMs' reasoning to be directly presented in the prompt, eliminating spatial hallucination of LLMs.

This study transforms the coordinates that describe the locations in the maze, such as \textit{(1,0)} into Node \textit{F} and \textit{(0,0)} into Node \textit{A}. The reachability between coordinates is converted into relationships between nodes. For example, if \textit{(1,0)} and \textit{(0,0)} are reachable, it is translated into \textit{(A,F)}, indicating a direct relationship between Node \textit{A} and Node \textit{F}. As a result, we used the letters \textit{A} and \textit{F} to represent nodes instead of numbers or coordinates. Our preliminary experiments indicate that LLMs exhibit generalization toward coordinates,  perceiving \textit{(1,0)} and \textit{(1,1)},  favoring movement toward \textit{(1,1)}, and leading to dead ends. By employing character representations, we can reduce node similarity and focus on relationships rather than node similarity.

\begin{figure}[!ht]
\includegraphics[width=\textwidth]{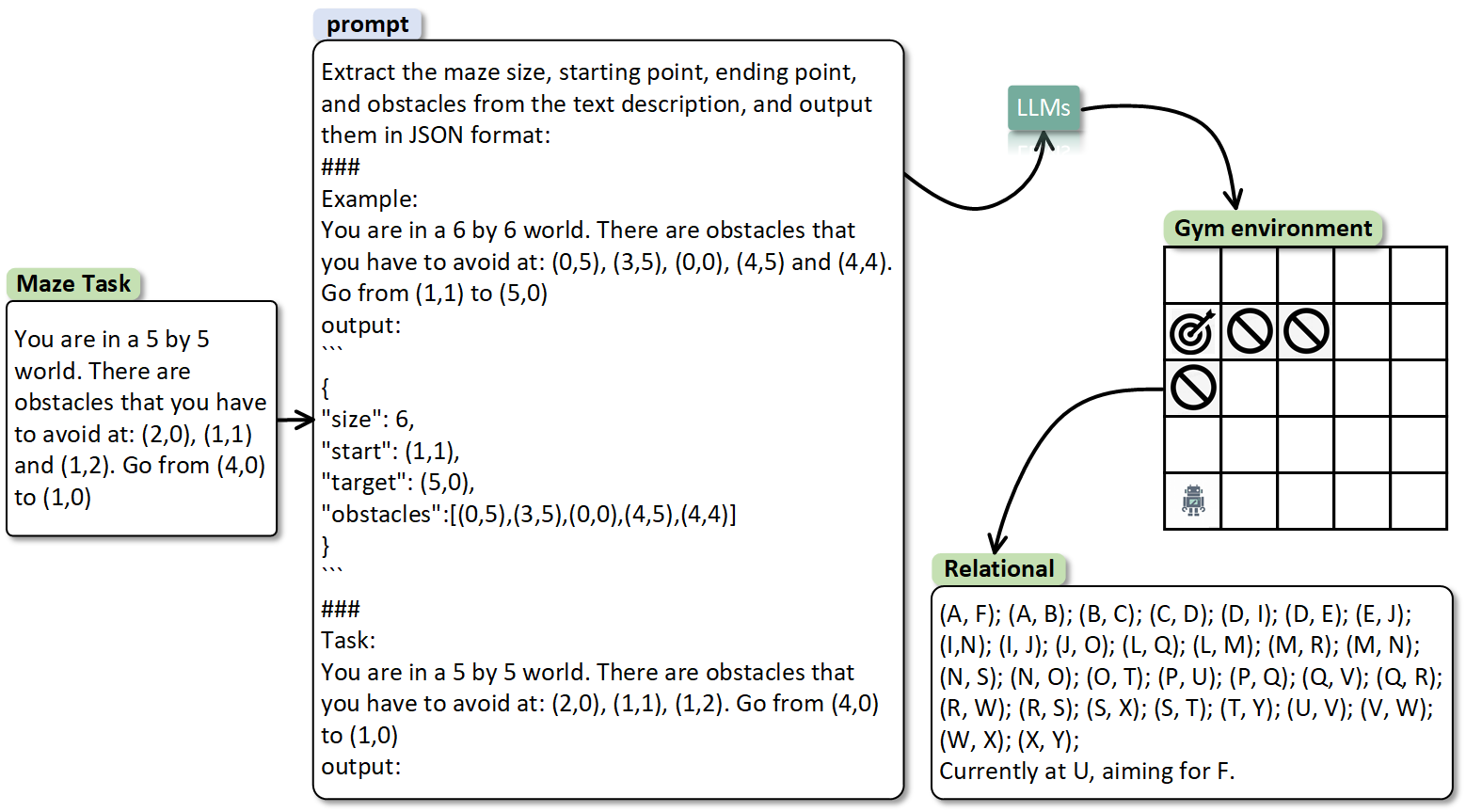}
\caption{This module can process any maze map description and convert it into a relational network.} \label{fig:S2R}
\end{figure}

Figure~\ref{fig:S2R} illustrates the process of Spatial-to-Relational Transformation. Initially, we transform the generically described maze into a structured representation using LLMs. In addition, we employed a \textit{JSON} format to represent maze information, including maze size, start and end points, and obstacle coordinates. Through this process, we can effectively extract structured information by instructing LLMs to output in JSON format. For this purpose, we employ a one-shot exemplar prompt. Subsequently, we leverage the OpenAI Gym~\cite{Brockman} environment to transform the structured maze into an interactive simulation environment, encompassing essential functions such as action execution, state updates, and reward calculations. Finally, we convert the state output from Gym into text, which includes information about the node relationship network.

\subsection{Curriculum Q-Learning}
\paragraph{Reverse Curriculum Generator with LLMs}
We used reverse curriculum learning (RCL)~\cite{Florensa} for LLMs inference, generating curricula from easy to complex tasks based on the prompt engineering of LLMs. The \textit{RCL} begins by starting from a state close to the goal using random walks to find a reachable initial state \textit{X1}, which is a simplified task. Then,  based on \textit{X1}, a more challenging initial state, X2, is generated, which continues iteratively. Therefore, the agent must learn according to the curriculum difficulty and transfer the experience from simple tasks to complex ones, thereby enhancing the efficiency of policy learning.

\begin{figure}[!ht]
\includegraphics[width=\textwidth]{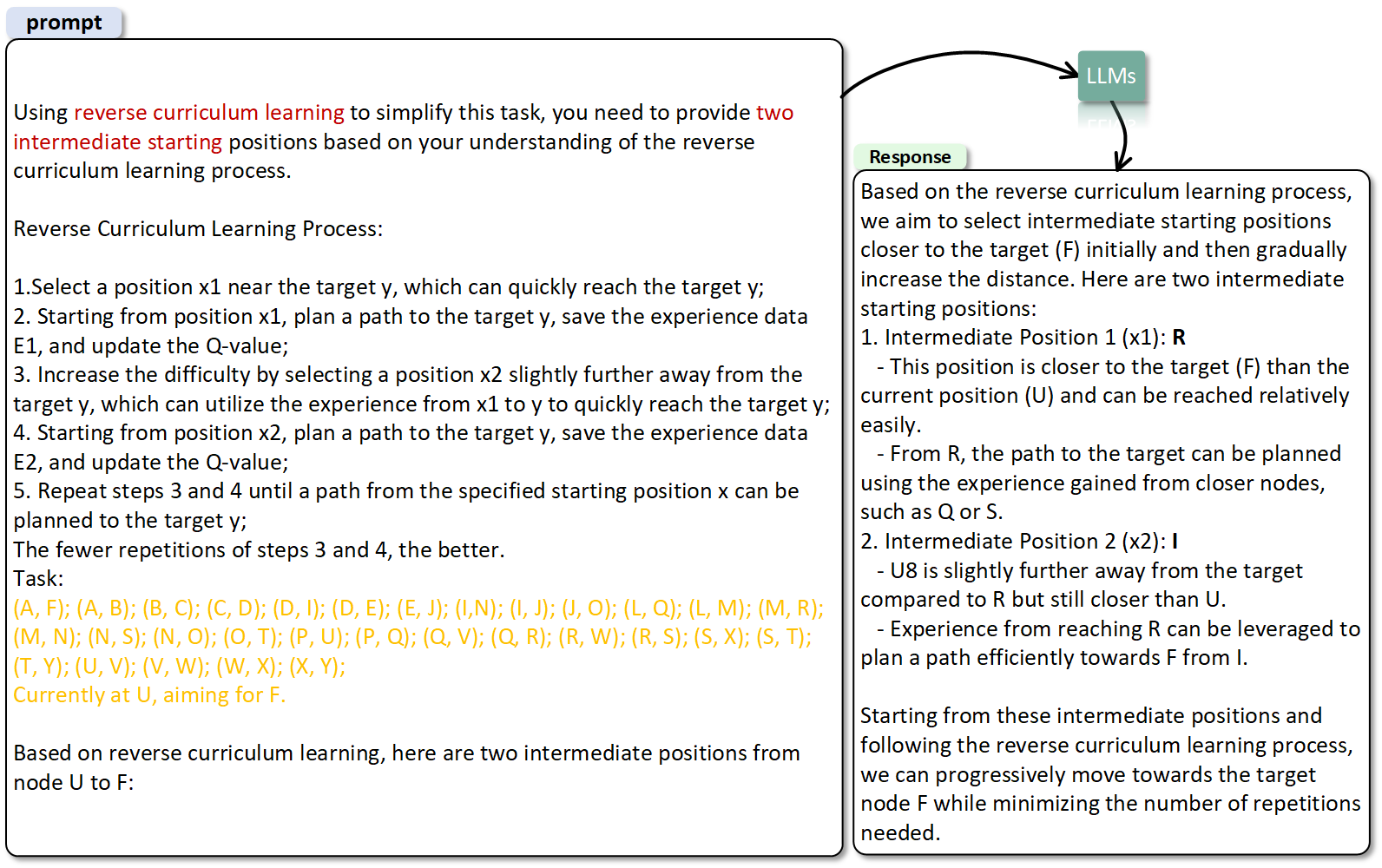}
\caption{Generate curriculums by LLMs.} \label{fig:CL}
\end{figure}

We extend this approach to the LLMs' prompts by describing the reverse curriculum generation process in natural language, allowing the LLMs to autonomously determine the simplified starting point, shown in Figure~\ref{fig:CL}. We incorporate hand-crafted curricula into S2RCQL to obtain a better curriculum, which proves beneficial for handling highly complex mazes. We performed experiments to thoroughly compare the quality of the two types of curricula generation and their impact on the algorithm.

\paragraph{Curriculum Q-learning}
We used LLMs or hand-crafted method courses to design a staged Q-learning optimization. By starting from a given initial point, we used the Q-learning algorithm to update the Q-table and store the experience data tuple $(s,a,r,s',q)$ in the experience replay buffer, where $s$ represents the current node; $a$ denotes the action taken by the agent; $r$ is the reward received; $s'$ is the node after the move, and $q$ represents the state-action value function of that sample. Specifically, we set the reward $r=-1$ for each step and $r=30$ upon reaching the goal, encouraging the agent to search the shortest path.

Equation~\ref{eq:ql} presents the action sampling function of Q-learning based on LLMs, using the $\epsilon-greedy$ algorithm. Compared with traditional Q-learning, we replace the random exploration component with LLMs, leveraging their prior knowledge to eliminate the agent's exploration cost, where $a_t$ indicates the action that should be taken at the current state, $s_t$ is the current state,  $q\_table$ is the state-action value table, and $argmax_aQ$ represents the action with the largest Q value. $p \in (0,1)$ is a random number sampled at each step.
\begin{equation} \label{eq:ql}
a_t =
\begin{cases}
LLMs(s_t, q\_table) & if\quad p < (1 - \epsilon) \\
argmax_aQ(s_t, a) & otherwise
\end{cases}
\end{equation}

\begin{algorithm}
    \renewcommand{\algorithmicrequire}{\textbf{Input:}}
    \renewcommand{\algorithmicensure}{\textbf{Output:}}
    \caption{S2RCQL} 
    \label{s2rcql}
    \begin{algorithmic}
        \REQUIRE $Maze\ text$
        \STATE env\_maze = LLM(Maze\ text)
        \STATE $C_1, C_2 = LLM(env\_maze)$
        \FOR{$C$ in $[C_1, C_2, start]$}
        \WHILE{$not\ success$}
        \STATE $success = Q\_learning(C, target, Exps, Q_{table})$
        \ENDWHILE
        \ENDFOR
        \ENSURE $Q_{table}$
    \end{algorithmic} 
\end{algorithm}

\paragraph{Pseudocode of Curriculum Q-learning}
Algorithm \ref{s2rcql} describes the pseudocode of \textbf{S2RCQL}. This study first inputs the general description of the maze. Then, we translate the description into a Gym environment based on LLMs and Python. LLMs generate the two courses ($C_1$ and $C_2$), representing two intermediate points in the maze path, which are easy to solve. Next, we use Q-learning with LLMs to plan the path from the starting point $C$ to the target point. This process is iterative until the path from the starting point $C$ is successfully solved. Finally, the algorithm outputs the history of path planning and $Q_{table}$,  representing the path plan policy.

\section{Experiments}
\subsection{Experiment Settings}
We performed numerous maze experiments based on Baidu's LLM \textbf{ERNIE-Bot 4.0}. To verify the inhibitory effect of S2RQL on hallucination, we compare to algorithms based on prompt engineering included \textit{CoT}, \textit{ToT}, \textit{React}, \textit{Q-learning}, and \textit{Rememberer}. Mazes of varying sizes were designed for the experiments, including 30 with $5\times5$ mazes, 20 with $7\times7$ mazes, and 10 with $10\times10$ mazes. The baseline \textbf{Prompt Engineering} are as follows: \textbf{Naive}, \textbf{CoT}~\cite{Wei}, \textbf{ToT}~\cite{Yao}, \textbf{ReAct}~\cite{Aghzal}, \textbf{Rememberer}~\cite{Zhang D}.

We evaluate the effectiveness of LLMs in maze planning by two indicators: 

(1) \textbf{Success Rate($\%$)}: The ratio of the successful attempts by LLM to reach the target compared with the total number of runs. This metric emphasizes its ability to reach the target, which should not require the shortest path. The formal definition is $Success\ Rate = \frac{\mathbb{N}_{suc}}{\mathbb{N}_{all}}$, where $\mathbb{N}_{suc}$ represents the number of successful attempts, and $\mathbb{N}_{all}$ represents the total number of runs.

(2) \textbf{Optimality Rate($\%$)}: The proportion of getting the shortest path in successful cases is defined as $Optimality\ Rate = \frac{\mathbb{N}_{opt}}{\mathbb{N}_{suc}}$, where $\mathbb{N}_{opt}$ represents the number of optimal cases. Specifically, many multiple shortest paths are found in a MAZE. However, the length of the shortest path is unique as long as the length of the resulting path reaches the minimum value.

\subsection{Main Results}
\begin{table}[!ht]
\caption{The results for each model are in all mazes, where $(n)$ represents training the model by $n$ episodes. The best results are highlighted in \textbf{Bold}, and the best baseline models are underlined}.\label{tab1}
\begin{center}
\renewcommand{\arraystretch}{1.2}
\begin{tabular}{ccccccc}
\toprule
\multirow{2.5}{*}{Method} &
\multicolumn{2}{c}{$5\times 5$}&
\multicolumn{2}{c}{$7\times 7$}&
\multicolumn{2}{c}{$10\times 10$}  \\
\cmidrule(lr){2-3}\cmidrule(lr){4-5}\cmidrule(lr){6-7}
  &Success &Optimality 
  &Success &Optimality 
  &Success &Optimality \\
\midrule
naive prompt&11.4\%&10.8\%&10.3\%&12.5\%&9.1\%&8.9\% \\
CoT~\cite{Wei}&15.0\%&14.5\%&14.9\%&13.5\%&10.5\%&10.1\% \\
ToT~\cite{Yao}&17.1\%&13.8\%&16.6\%&13.1\%&10.3\%&12.9\% \\ 
ReAct~\cite{Aghzal}&17.4\%&22.8\%&16.1\%&21.6\%&15.4\%&20.7\% \\
$Rememberer_{(30)}$~\cite{Zhang D}&\underline{45.1\%}&\underline{{50.8\%}}&\underline{40.2\%}&\underline{44.2\%}&\underline{34.8\%}&\underline{35.7\%} \\
\midrule
$S2RCQL_{(30)}$&\textbf{85.6\%}&\textbf{73.8\%}&\textbf{73.4\%}&\textbf{69.6\%}&\textbf{64.7\%}&\textbf{65.7\%} \\

\bottomrule
\end{tabular}
\end{center}
\end{table}

\begin{figure}[!ht]
\centering
\subfigure[in 5$\times$5 maze]{
\includegraphics[scale=0.3,width=0.3\textwidth]{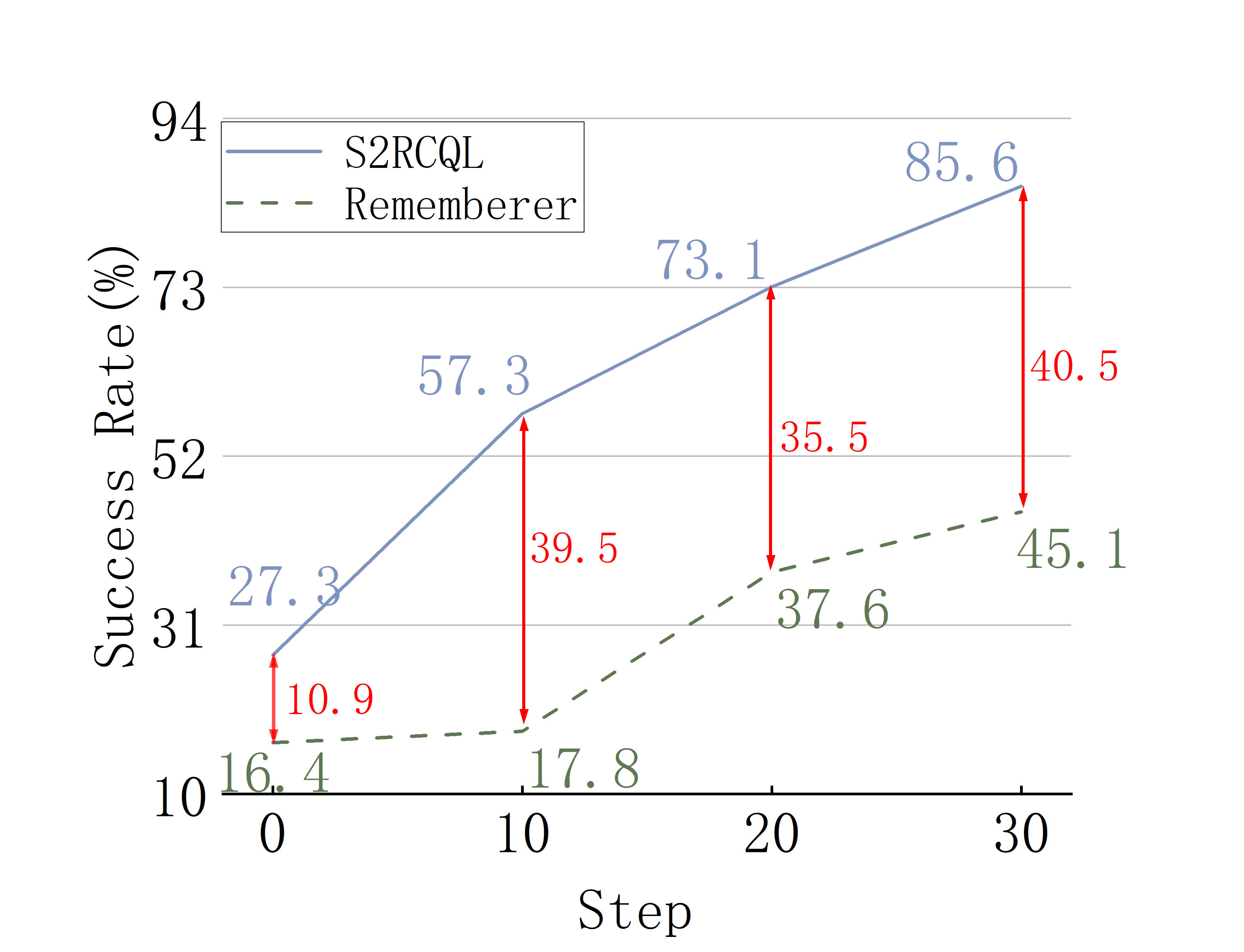}
}
\subfigure[in 7$\times$7 maze]{
\includegraphics[scale=0.3,width=0.3\textwidth]{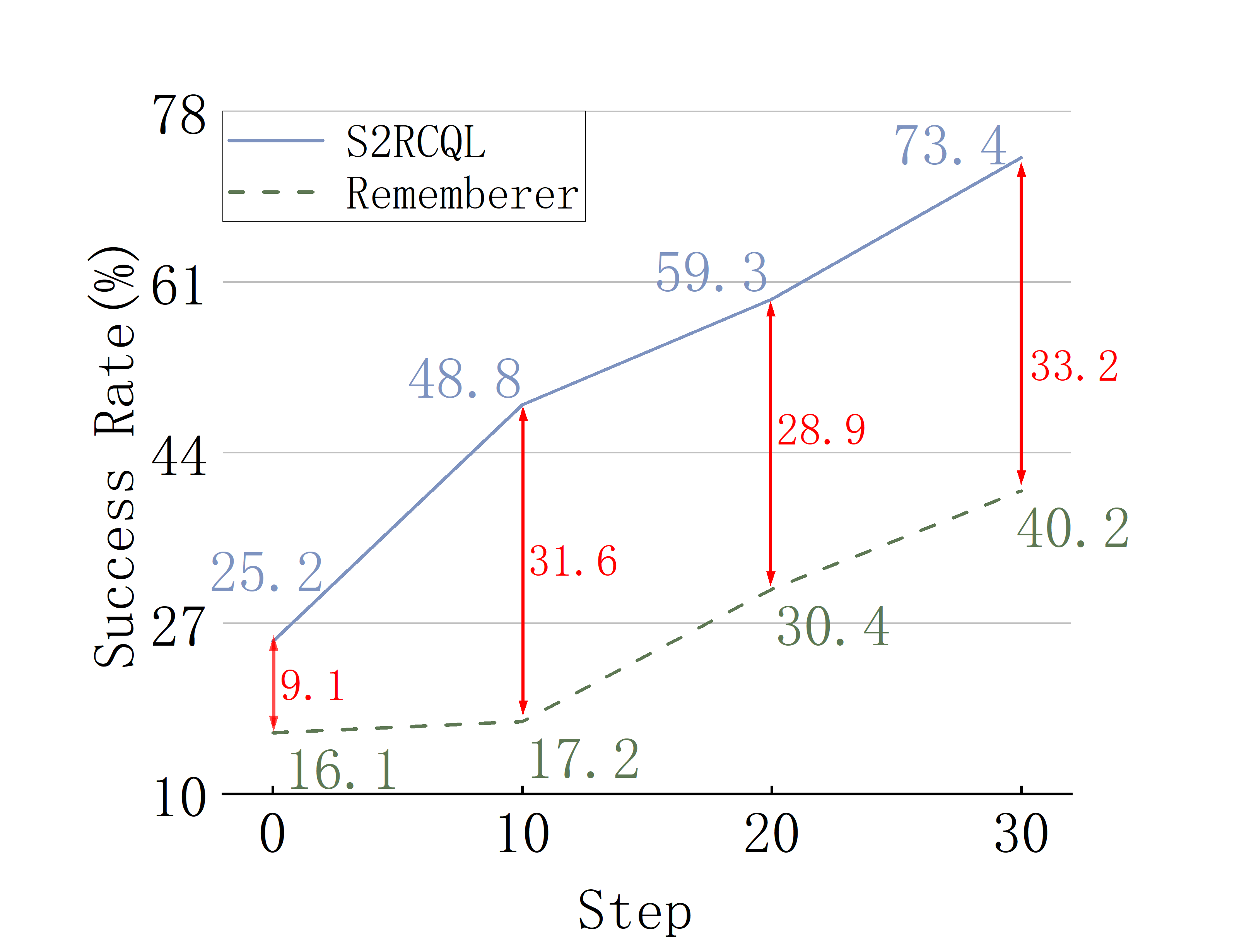}
}
\subfigure[in 10$\times$10 maze]{
\includegraphics[scale=0.3,width=0.3\textwidth]{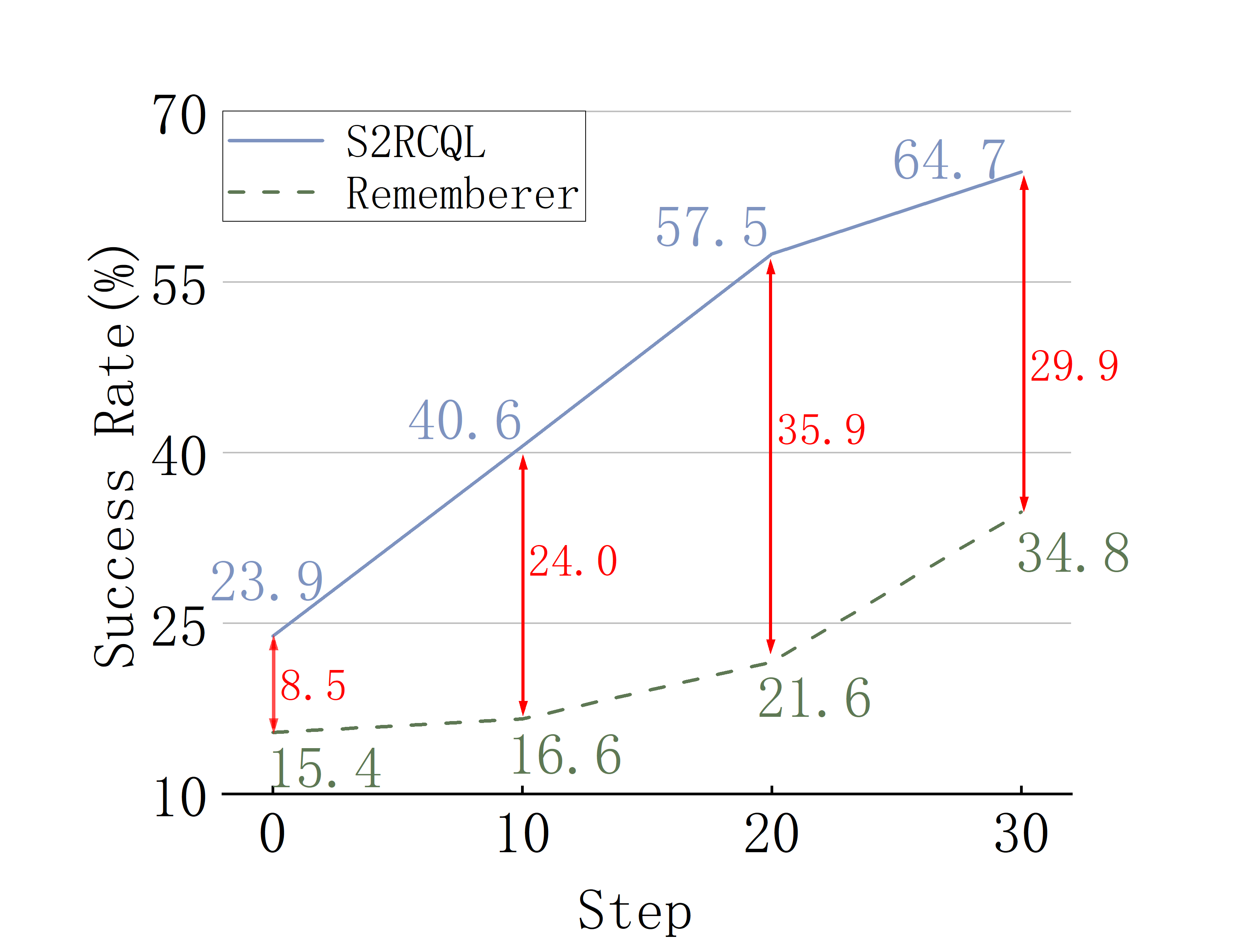}
}
\subfigure[in 5$\times$5 maze]{
\includegraphics[scale=0.3,width=0.3\textwidth]{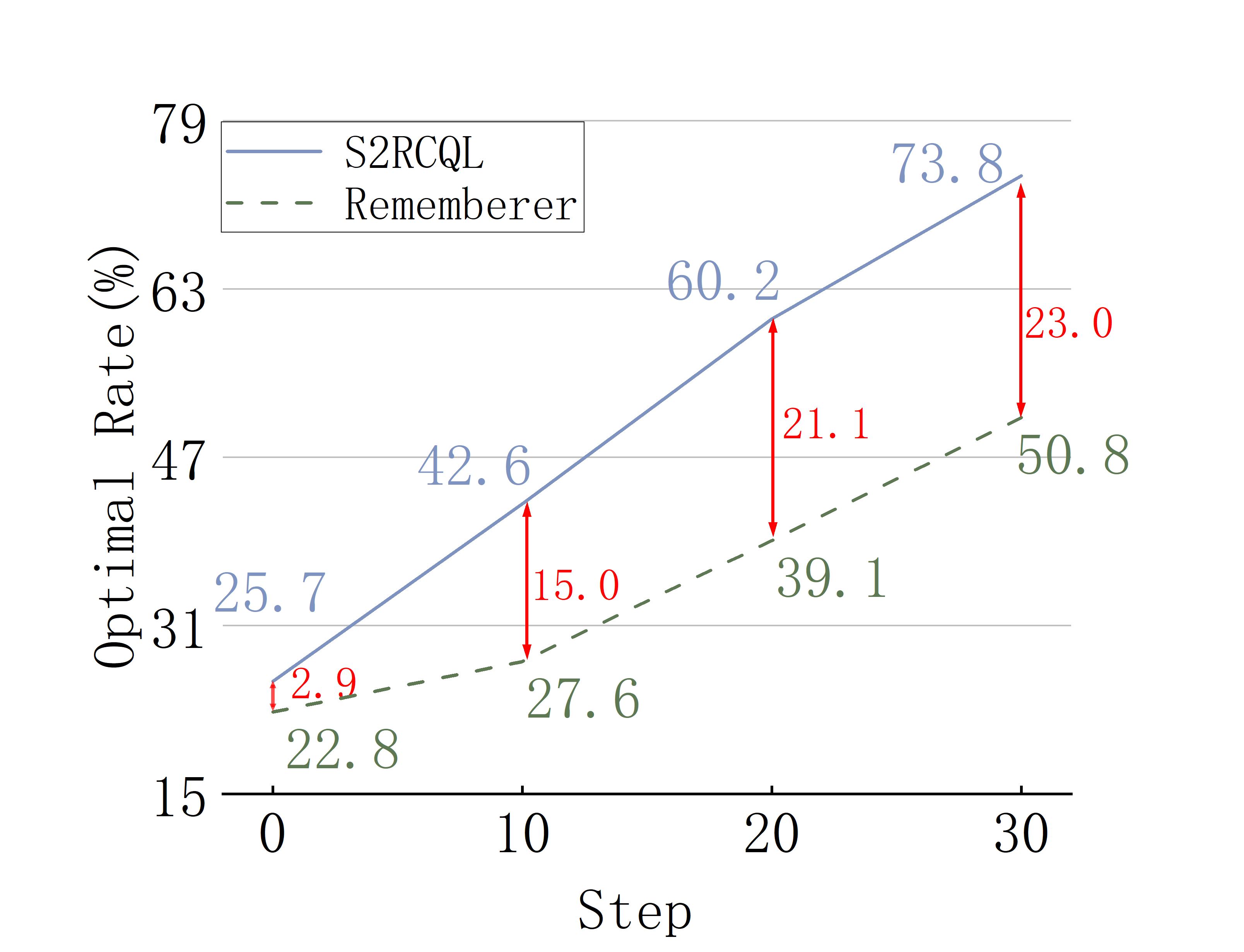}
}
\subfigure[in 7$\times$7 maze]{
\includegraphics[scale=0.3,width=0.3\textwidth]{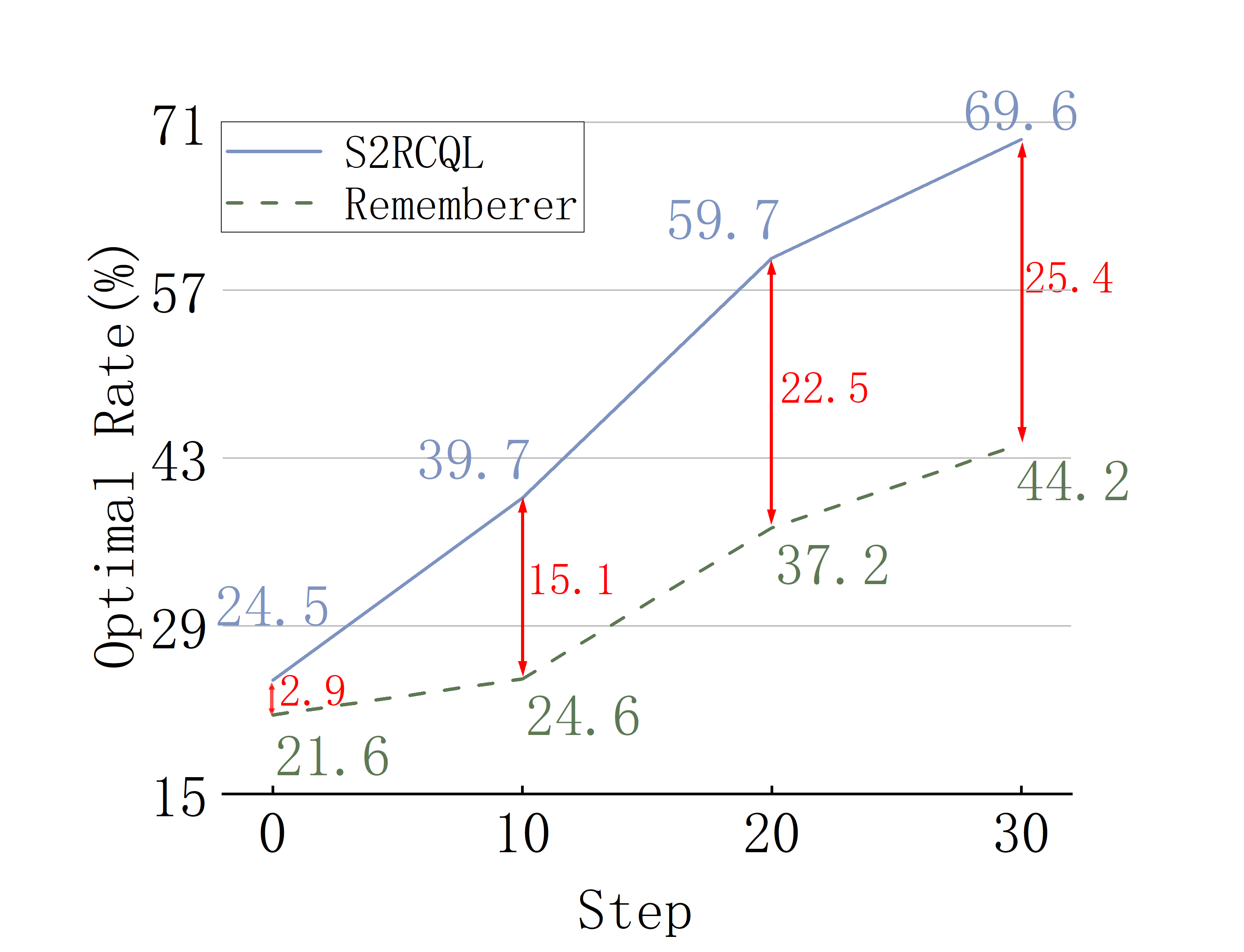}
}
\subfigure[in 10$\times$10 maze]{
\includegraphics[scale=0.3,width=0.3\textwidth]{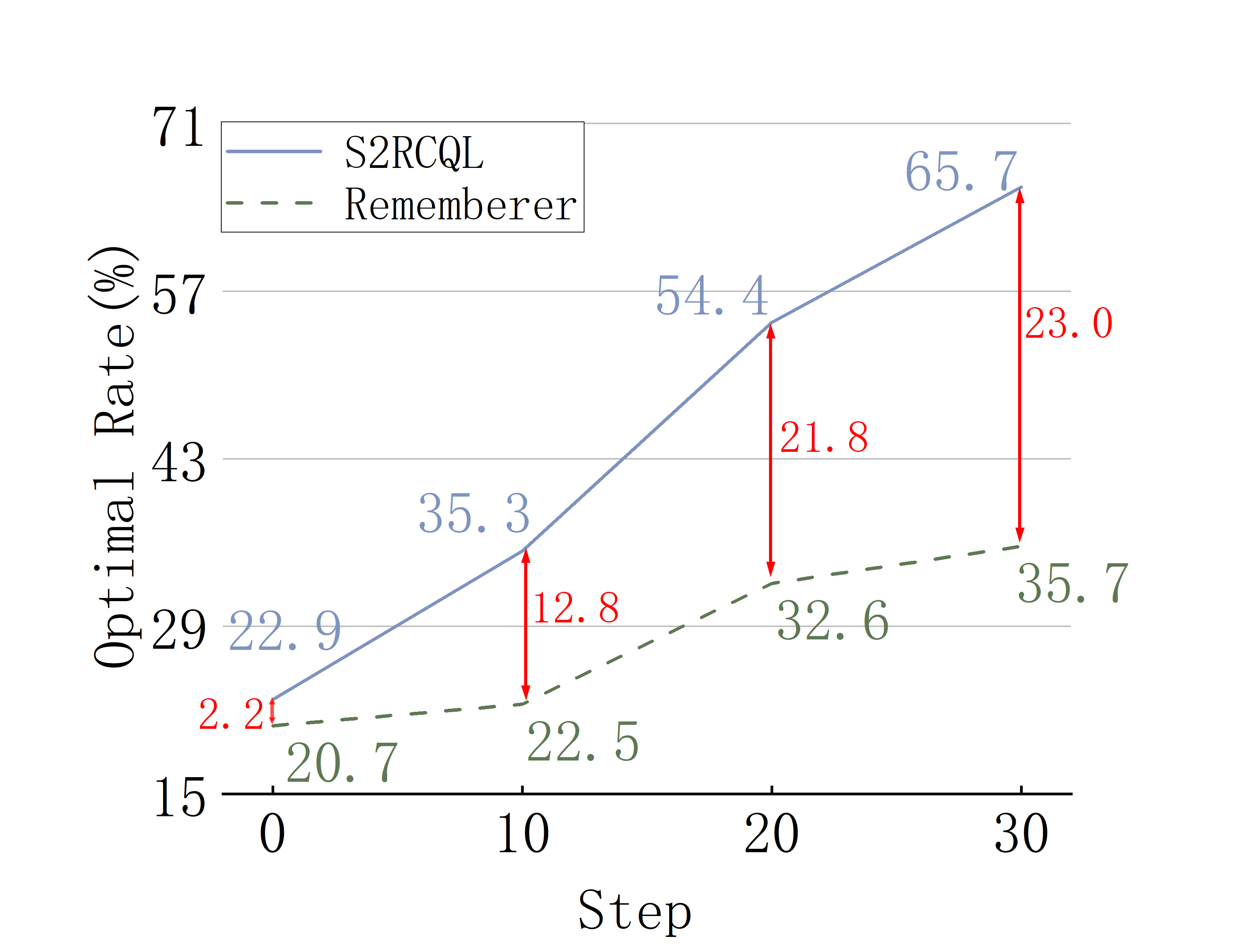}
}
\caption{Comparison of the performance of \textbf{S2RCQL} and \textbf{Rememberer}.} \label{fig:cmp}
\end{figure}

Comparison of the performance of S2RCQL and Rememberer algorithms in mazes of different sizes. 
Table~\ref{tab1} displays the main results of each method based on the \textit{Success} and \textit{Optimal} indicators. From the experimental results,  our S2RCQL outperforms the Rememberer algorithm by 25\%--40\% in terms of \textit{Success} Rate and 23\%--30\% in terms of \textit{Optimality} Rate. We also compare to Rememberer along with the training episode, which is shown in Figure~\ref{fig:cmp}. Furthermore, as the maze size increases, the \textit{Success} Rate and \textit{Optimality} Rate gradually decline because the maze size determines the search space.

\begin{table}[!ht]
\caption{Enhancement of various prompt engineering with the aid of the S2R module.}\label{tab2}
\begin{center}
\renewcommand{\arraystretch}{1.2}
{\begin{tabular}{ccccccc}
\toprule
\multirow{2.5}{*}{Method} &
\multicolumn{2}{c}{$5\times 5$}&
\multicolumn{2}{c}{$7\times 7$}&
\multicolumn{2}{c}{$10\times 10$}  \\
\cmidrule(lr){2-3}\cmidrule(lr){4-5}\cmidrule(lr){6-7}
  &Success &Optimality 
  &Success &Optimality 
  &Success &Optimality \\
\midrule
naive prompt&11.4\%&10.8\%&10.3\%&12.5\%&9.1\%&8.9\% \\
naive prompt w/ S2R&25.3\%&27.1\%&13.3\%&23.9\%&10.1\%&18.7\% \\
\midrule
CoT&15.0\%&14.5\%&14.9\%&13.5\%&10.5\%&10.1\% \\
CoT w/ S2R&30.9\%&33.5\%&20.9\%&23.1\%&13.5\%&19.4\% \\
\midrule
ToT&17.1\%&13.8\%&16.6\%&13.1\%&10.3\%&12.9\% \\ 
ToT w/ S2R&33.4\%&38.8\%&24.6\%&27.5\%&14.5\%&20.9\% \\
\midrule
ReAct&17.4\%&22.8\%&16.1\%&21.6\%&15.4\%&20.7\% \\
ReAct w/ S2R&35.7\%&40.8\%&27.6\%&30.3\%&19.7\%&23.6\% \\
\midrule
$Rememberer_{(30)}$&45.1\%&50.8\%&40.2\%&44.2\%&34.8\%&35.7\% \\
$Rememberer\ w/\ S2R_{(30)}$&61.9\%&59.6\%&55.6\%&47.9\%&47.5\%&41.1\% \\
\bottomrule
\end{tabular}}
\end{center}
\end{table}

\begin{figure}[!ht]
\centering
\begin{minipage}{\textwidth}
  \centering
  \subfigure[in 5$\times$5 maze]{
  \includegraphics[scale=0.3]{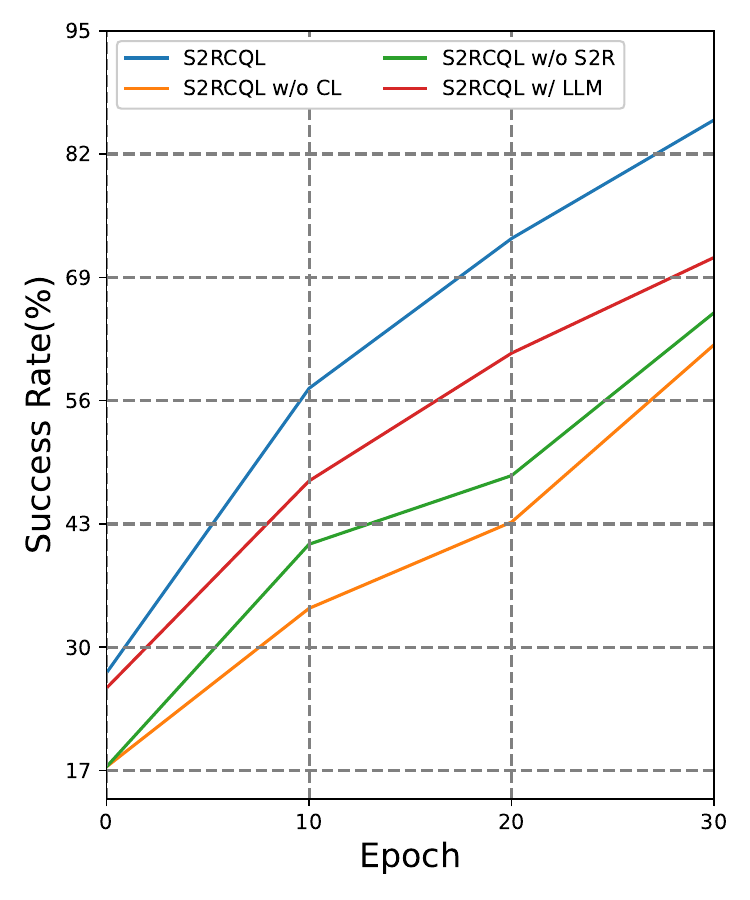}
  }
  \subfigure[in 7$\times$7 maze]{
  \includegraphics[scale=0.3]{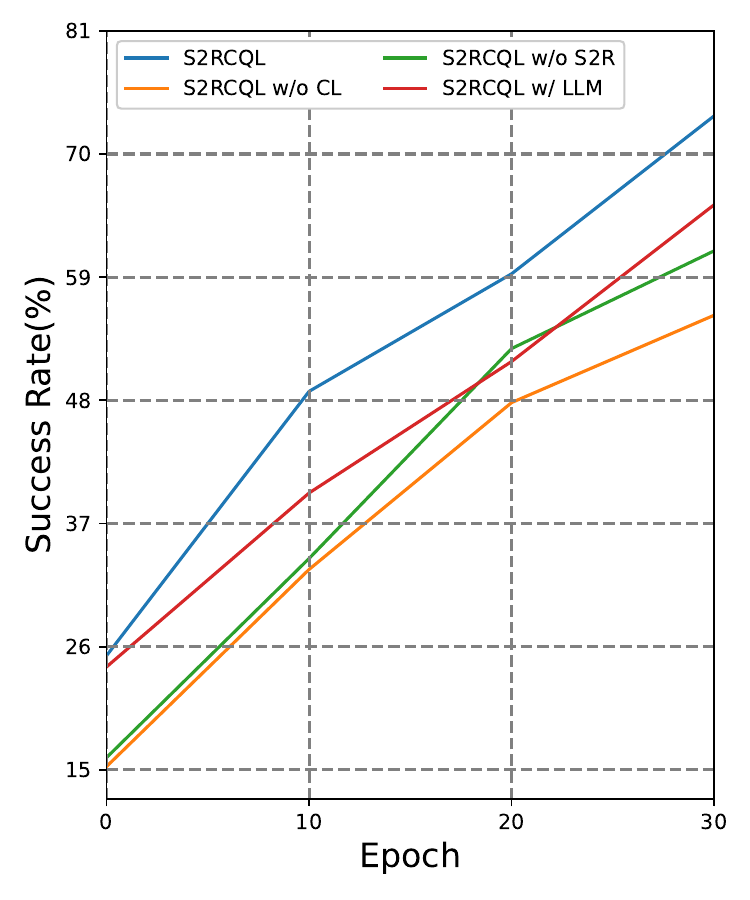}
  }
  \subfigure[in 10$\times$10 maze]{
  \includegraphics[scale=0.3]{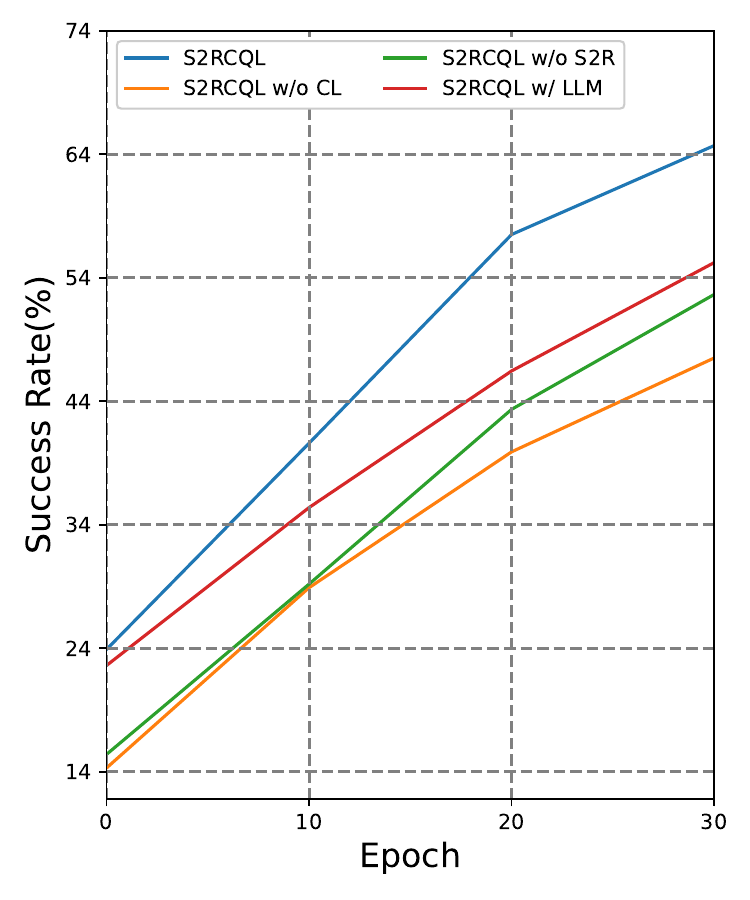}
  }
\end{minipage}

\vspace{0.5cm} 

\begin{minipage}{\textwidth}
  \centering
  \subfigure[in 5$\times$5 maze]{
  \includegraphics[scale=0.3]{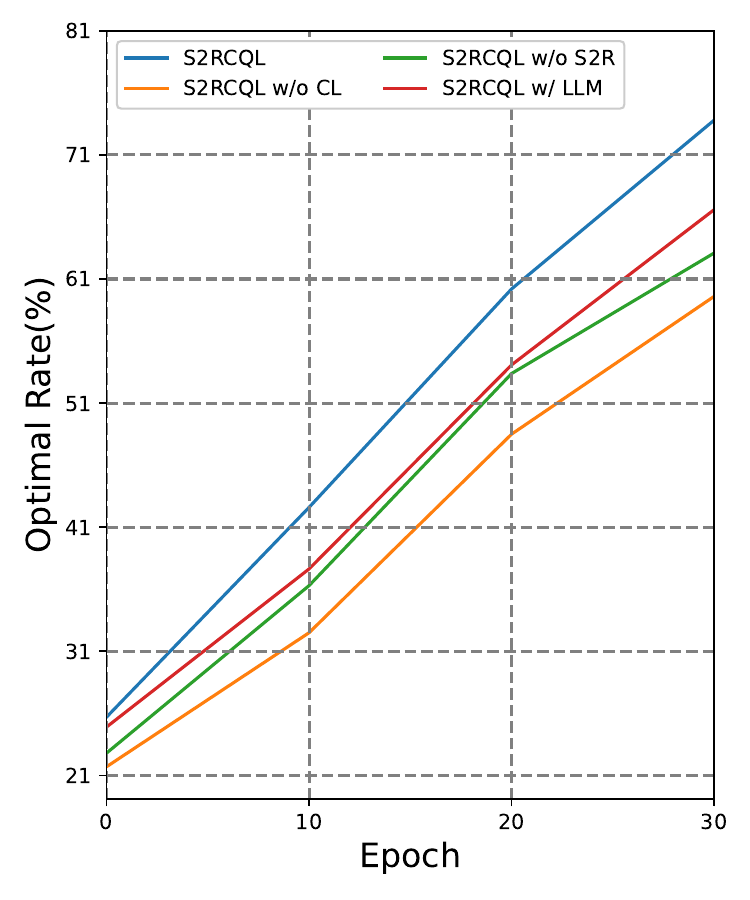}
  }
  \subfigure[in 7$\times$7 maze]{
  \includegraphics[scale=0.3]{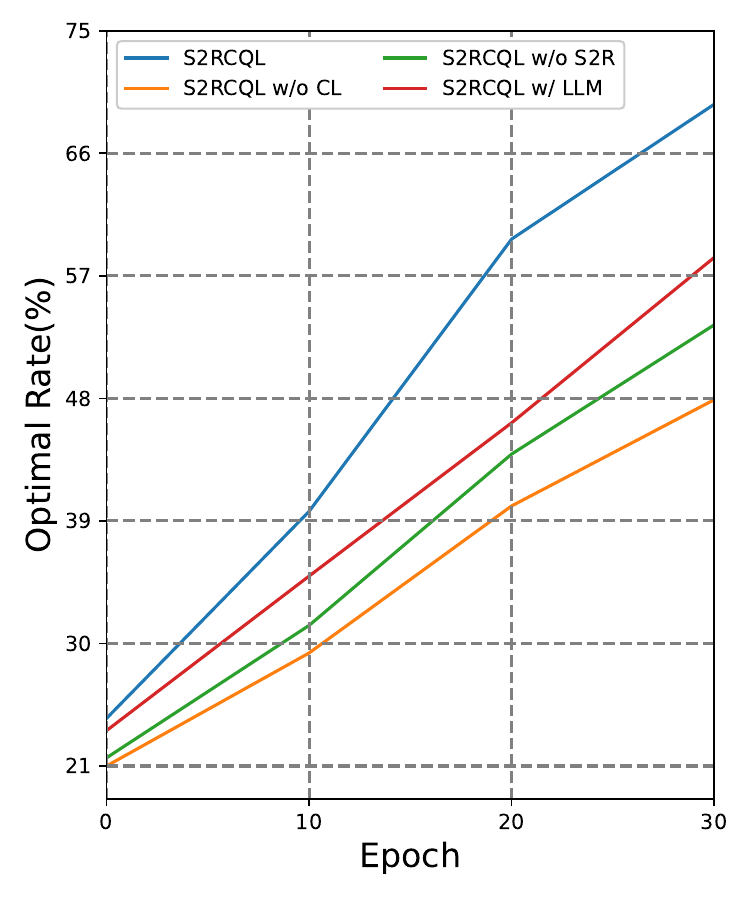}
  }
  \subfigure[in 10$\times$10 maze]{
  \includegraphics[scale=0.3]{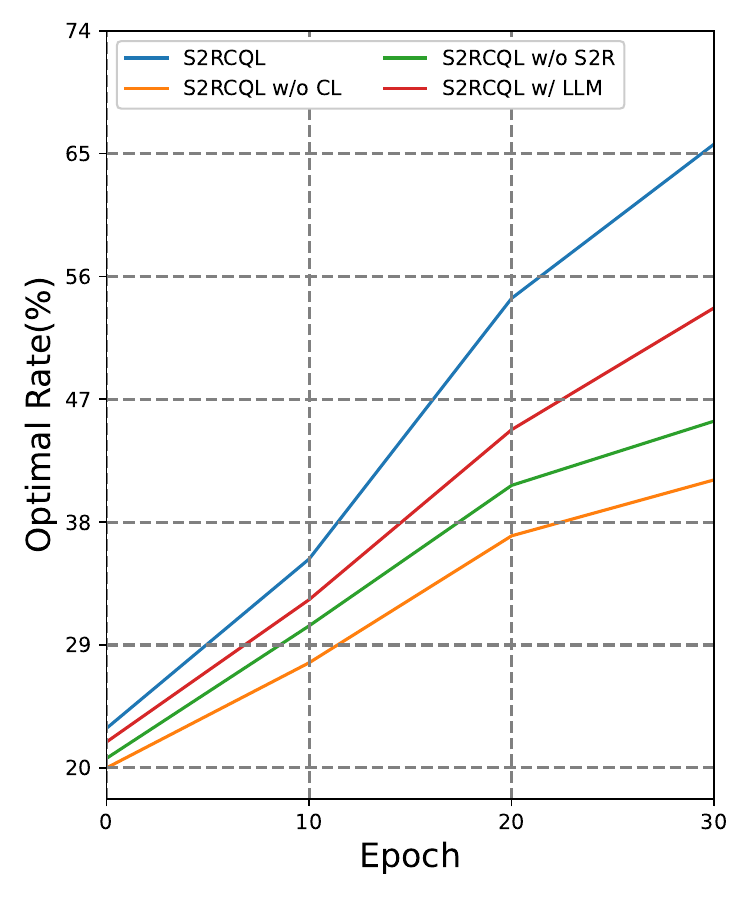}
  }
\end{minipage}

\caption{Comparison of the effectiveness of the \textbf{S2RCQL} algorithm without course or \textbf{S2R} and under different curriculum generation schemes.} \label{fig:abs}
\end{figure}

\subsection{Ablation study}

We applied the S2R module to various prompt engineering for verifying the generality and effectiveness of S2R. The experimental results are presented in Table\textbf{~\ref{tab2}. }The results demonstrate that our S2R significantly improves both Success and Optimality rates across various algorithms by mitigating the LLM's spatial hallucination. We remove the \textbf{S2R} module from \textbf{S2RCQL}. The results indicated a decline of approximately 15\% in both the \textit{Success} and \textit{Optimality} rates, shown in Figure~\ref{fig:abs}. We removed the \textbf{CL} module from \textbf{S2RCQL}. The results indicate that the \textit{Success} and \textit{Optimality} rates decrease by approximately 20\% when there is an absence of curriculum learning. As the maze size increases, the impact of \textbf{CL} becomes more significant. Furthermore, we conducted a comparison between hand-crafted and LLM-generated reverse curricula, as shown in Figure~\ref{fig:abs}. The curriculum generated by LLMs demonstrates an improvement of approximately 10\% compared with algorithms without curriculum learning.

\begin{figure}[H]
\centering
\subfigure[Errors caused by various prompt engineering over long distances.]{
\includegraphics[scale=0.5,width=4in]{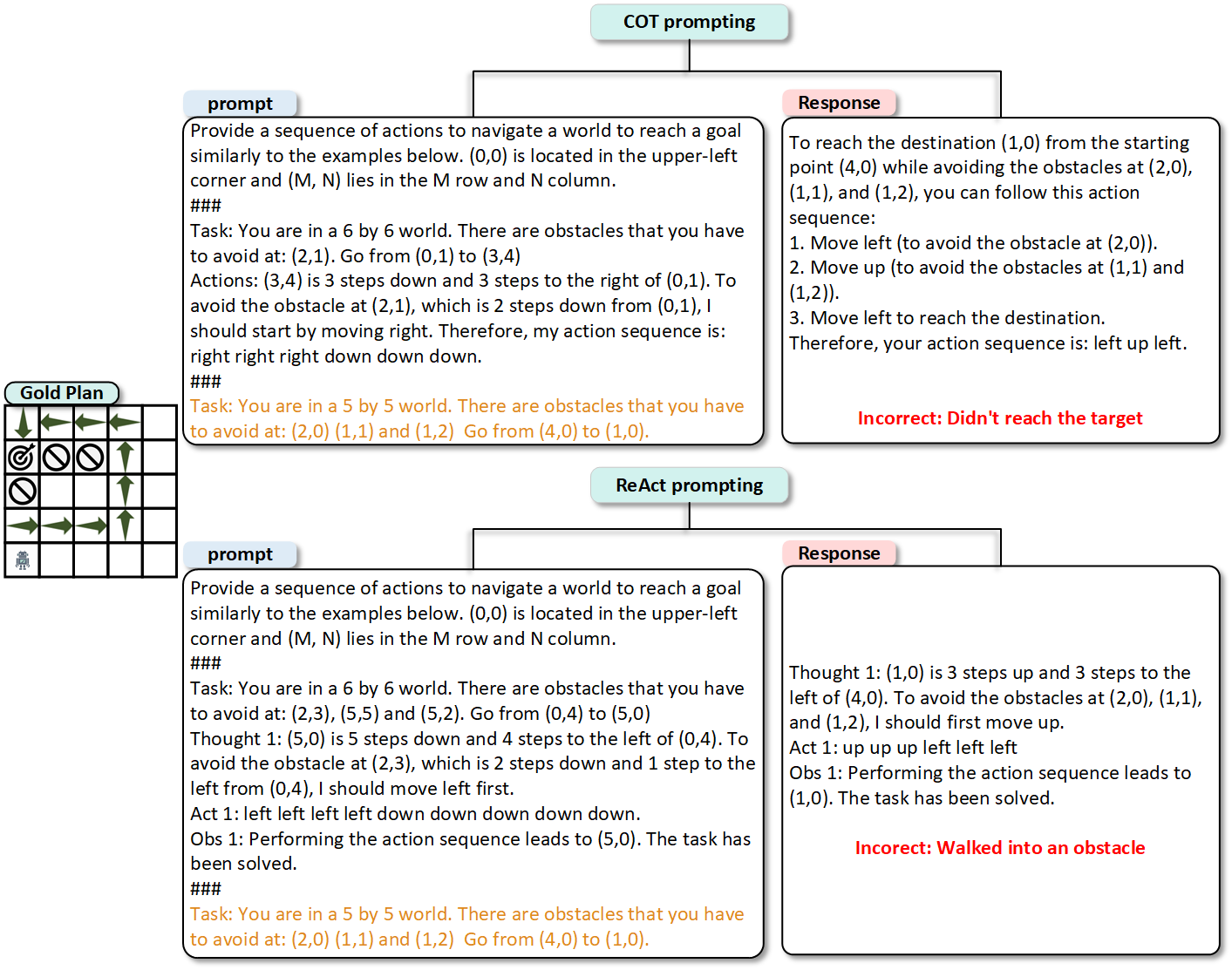} \label{graph_1}
}
\end{figure}

\addtocounter{figure}{0}

\begin{figure}[H]
\centering
\subfigure[Errors with prompting caused by various prompt engineering over short distances.]{
\includegraphics[width=\textwidth]{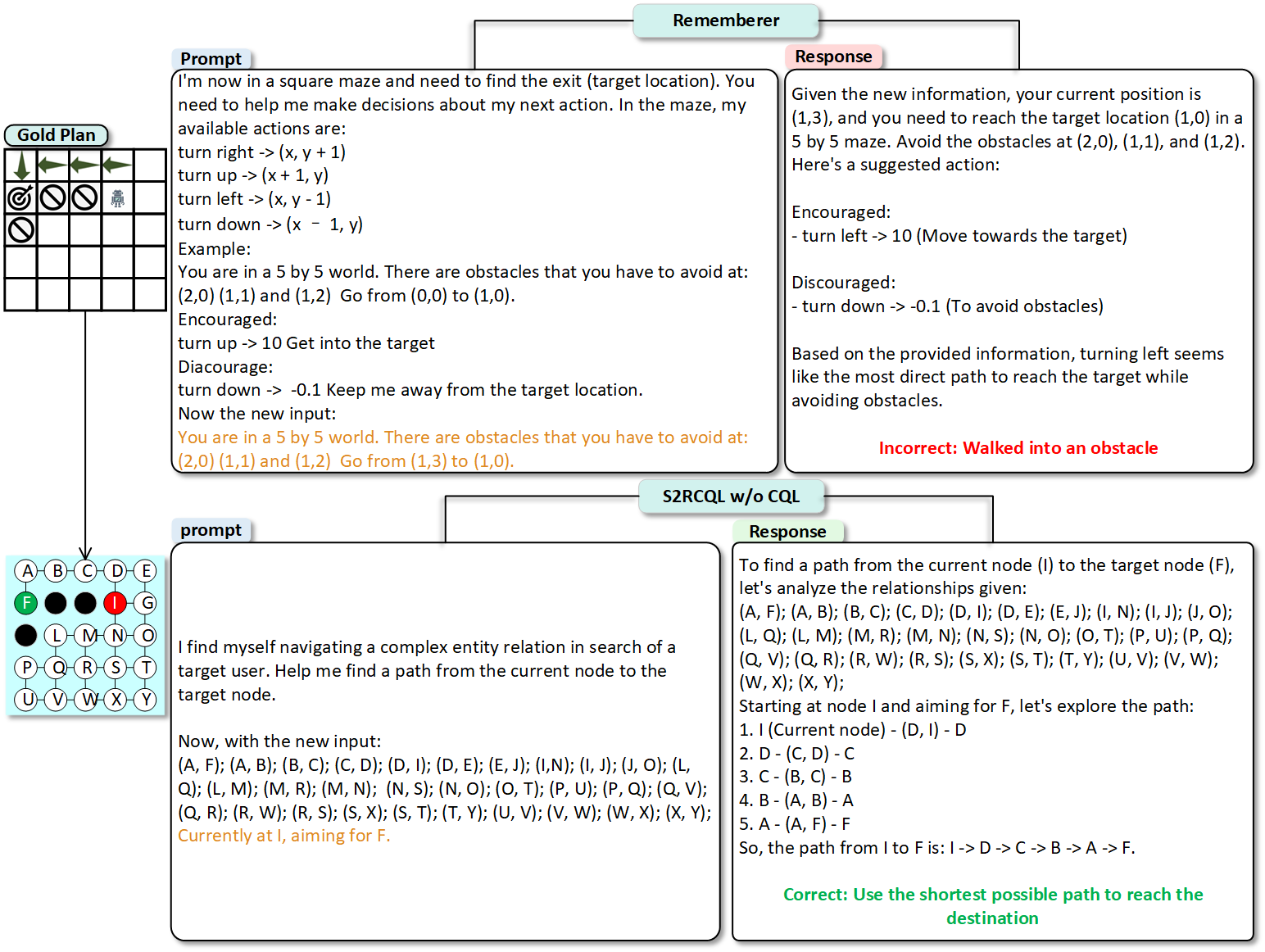} \label{graph_2}
}
\subfigure[Errors with Rememberer w/ S2R caused by prompt engineering over long distances.]{
\includegraphics[scale=0.5,width=4in]{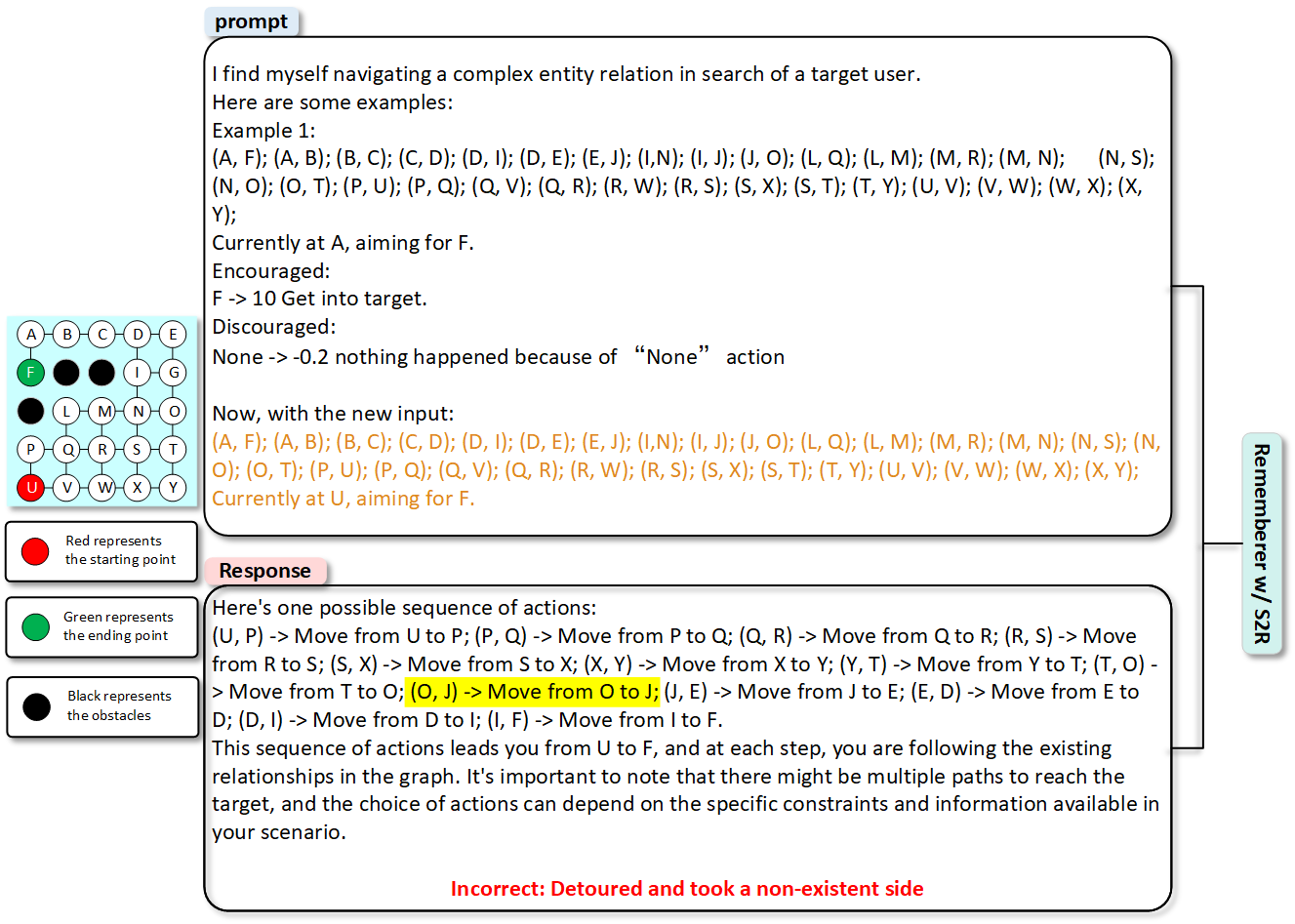} \label{graph_3}
}
\end{figure}

\addtocounter{figure}{-1} 

\begin{figure}[H]
\addtocounter{figure}{1} 
\centering
\subfigure[Corrections caused by S2RCQL prompt engineering over long distances.]{
\includegraphics[width=\textwidth]{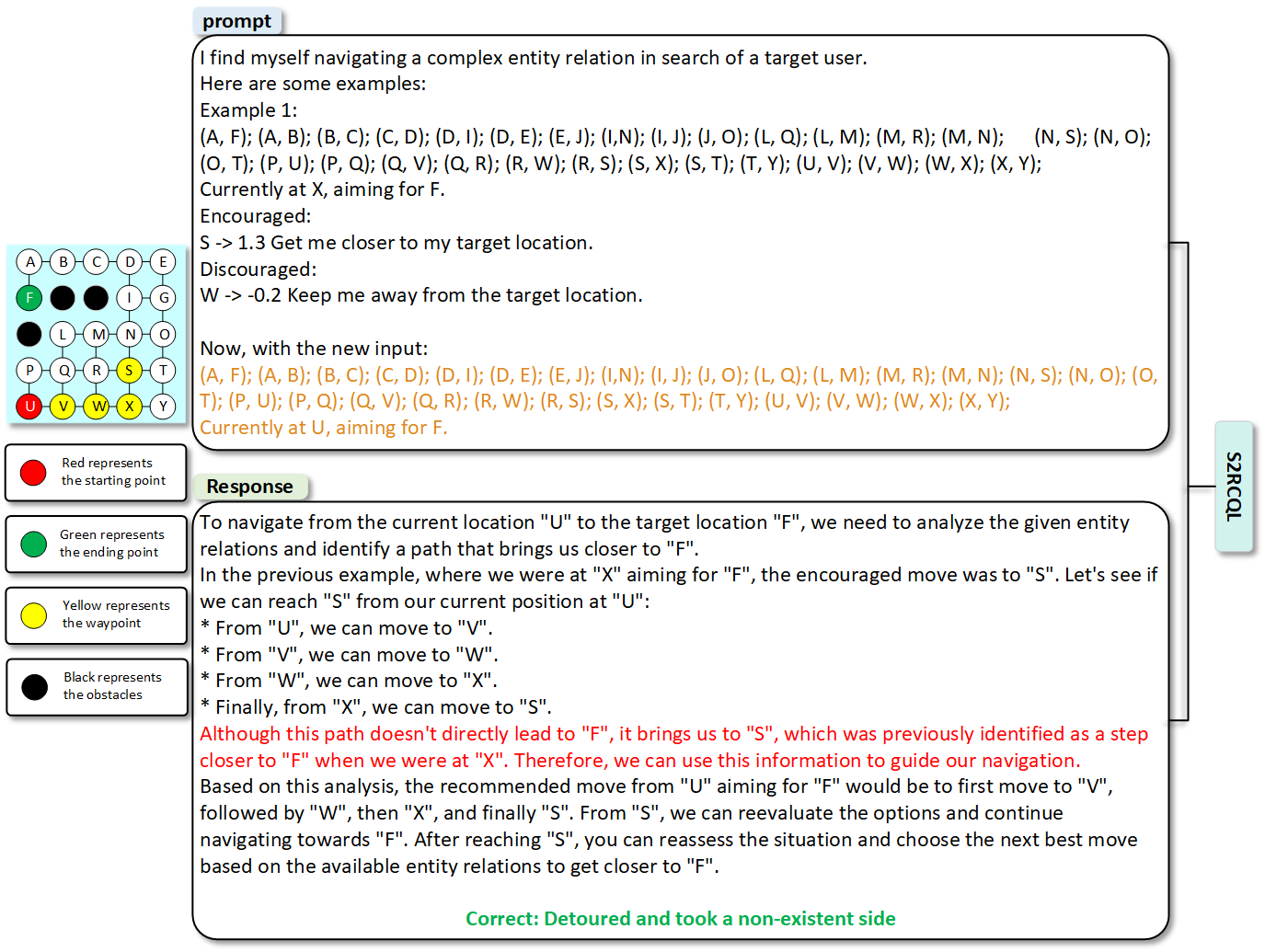} \label{graph_4}
}
\caption{Toy examples of the most common errors produced by each prompt engineering. We show the shortcomings of each method.}\label{fig:exp}
\end{figure}

\subsection{Toy Example}
 We conducted several case studies to intuitively demonstrate the execution process of LLMs, as  shown in Figure~\ref{fig:exp}:

(1) As demonstrated in Figure \ref{graph_1},  the agent reaches \textit{(2,1)} and \textit{(2,2)} in the prompt engineering. After that, due to spatial hallucination, the direction of action is lost, resulting in an unsolvable result. As a result, the agent directly faces the obstacle, and the prompt declares that the obstacle cannot be entered. This indicates context inconsistency hallucination in the Path Planning problem.

(2) As demonstrated in Figure \ref{graph_2}, when using the Rememberer model to navigate the maze, the shortest distance of this maze is small, and LLMs attempt to incorrectly navigate to position \textit{(1,2)} at the point of the obstacle. However, when we use entity relations, a simple prompt is required to navigate to the destination, proving that entity relations can enhance LLMs' space understanding and relieve the spatial hallucination in LLMs. In our experiment, we found that when navigating from the position \textit{(0,3)} to the target position \textit{(1,0)}, LLMs can get \textit{(1,0)} without error. 

(3) As shown in Figure \ref{graph_3}, we use Rememberer with S2R module. While the success rate is nearly 100\% for short-distance navigation, the performance is still poor for long-distance navigation. In long-term reasoning, the Rememberer still mistakenly assumes the existence of a relation \textit{(J, O)}. This shows that S2R alone is not sufficient to address LLMs' context inconsistency hallucination by long-term reasoning. As shown in Figure \ref{graph_4}, we propose the S2RCQL model to solve this problem. LLMs can accurately navigate to positions they have previously reached (such as point \textit{S} depicted in Figure \ref{graph_4}). This to some extent alleviates the context inconsistency hallucination. By introducing curriculum learning and the S2R module, S2RCQL can better utilize historical experience and find solutions.

\section{Conclusion}
This study proposes the \textbf{S2RCQL} algorithm to improve LLMs' path planning ability by alleviating spatial hallucinations and context inconsistency hallucinations in LLMs. Through S2R, we automatically convert the maze described by coordinates into an entity relation graph structure. The proposed \textbf{S2R} exhibits generality, and its application to various prompt engineering yields significant improvements. In addition, we design a reverse curriculum generation based on LLMs and a curriculum Q-learning algorithm that significantly improves the success rate and optimality rate of the maze path planning task. In future work, we will research how to enhance LLMs' reverse curriculum generation capabilities.

\bibliographystyle{unsrt}  


\begin{thebibliography}{1}

\bibitem{Imani}
Imani S, Du L, Shrivastava H. MathPrompter: Mathematical Reasoning using Large Language Models. InICLR 2023 Workshop on Trustworthy and Reliable Large-Scale Machine Learning Models 2023 Apr 16.

\bibitem{Dorbala}
Dorbala VS, Mullen Jr JF, Manocha D. 
\newblock Can an Embodied Agent Find Your “Cat-shaped Mug”? LLM-Based Zero-Shot Object Navigation.
\newblock IEEE Robotics and Automation Letters. 2023 Dec 25.

\bibitem{Piggott}
Piggott B, Patil S, Feng G, Odat I, Mukherjee R, Dharmalingam B, Liu A. Net-GPT: A LLM-Empowered Man-in-the-Middle Chatbot for Unmanned Aerial Vehicle. In2023 IEEE/ACM Symposium on Edge Computing (SEC) 2023 Dec 6 (pp. 287-293). IEEE.

\bibitem{Zhu}
Zhu X, Chen Y, Tian H, Tao C, Su W, Yang C, Huang G, Li B, Lu L, Wang X, Qiao Y. Ghost in the minecraft: Generally capable agents for open-world enviroments via large language models with text-based knowledge and memory. arXiv preprint arXiv:2305.17144. 2023 May 25.

\bibitem{Ye}
Ye S, Hwang H, Yang S, Yun H, Kim Y, Seo M. In-context instruction learning. arXiv e-prints. 2023 Feb:arXiv-2302.

\bibitem{Wei}
Wei J, Wang X, Schuurmans D, Bosma M, Xia F, Chi E, Le QV, Zhou D. 
\newblock Chain-of-thought prompting elicits reasoning in large language models. 
\newblock Advances in neural information processing systems. 2022 Dec 6;35:24824-37.

\bibitem{Chu}
Chu Z, Chen J, Chen Q, Yu W, He T, Wang H, Peng W, Liu M, Qin B, Liu T. A survey of chain of thought reasoning: Advances, frontiers and future. arXiv preprint arXiv:2309.15402. 2023 Sep 27.

\bibitem{Feng}
Feng G, Zhang B, Gu Y, Ye H, He D, Wang L. Towards revealing the mystery behind chain of thought: a theoretical perspective. Advances in Neural Information Processing Systems. 2024 Feb 13;36.

\bibitem{Yao}
Yao S, Yu D, Zhao J, Shafran I, Griffiths T, Cao Y, Narasimhan K. 
\newblock Tree of thoughts: Deliberate problem solving with large language models. 
\newblock Advances in Neural Information Processing Systems. 2024 Feb 13;36.

\bibitem{Besta}
Besta M, Blach N, Kubicek A, Gerstenberger R, Podstawski M, Gianinazzi L, Gajda J, Lehmann T, Niewiadomski H, Nyczyk P, Hoefler T. 
\newblock Graph of thoughts: Solving elaborate problems with large language models. 
\newblock In Proceedings of the AAAI Conference on Artificial Intelligence 2024 Mar 24 (Vol. 38, No. 16, pp. 17682-17690).

\bibitem{Yao Y}
Yao Y, Li Z, Zhao H. Beyond chain-of-thought, effective graph-of-thought reasoning in large language models. arXiv preprint arXiv:2305.16582. 2023 May 26.

\bibitem{Wang}
Wang X, Wei J, Schuurmans D, Le QV, Chi EH, Narang S, Chowdhery A, Zhou D. 
\newblock Self-Consistency Improves Chain of Thought Reasoning in Language Models. 
\newblock In The Eleventh International Conference on Learning Representations 2022 Sep 29.

\bibitem{Xiao Z}
Xiao Z, Zhang D, Wu Y, Xu L, Wang YJ, Han X, Fu X, Zhong T, Zeng J, Song M, Chen G. 
\newblock Chain-of-Experts: When LLMs Meet Complex Operations Research Problems. 
\newblock In The Twelfth International Conference on Learning Representations 2023 Oct 13.

\bibitem{Yao S}
Yao S, Zhao J, Yu D, Du N, Shafran I, Narasimhan K, Cao Y. 
\newblock ReAct: Synergizing Reasoning and Acting in Language Models. 
\newblock In International Conference on Learning Representations (ICLR) 2023 Jan.

\bibitem{Aghzal}
Aghzal M, Plaku E, Yao Z. 
\newblock Can large language models be good path planners? a benchmark and investigation on spatial-temporal reasoning. 
\newblock arXiv preprint arXiv:2310.03249. 2023 Oct 5.

\bibitem{Shinn}
Shinn N, Cassano F, Gopinath A, Narasimhan K, Yao S. 
\newblock Reflexion: Language agents with verbal reinforcement learning. 
\newblock Advances in Neural Information Processing Systems. 2024 Feb 13;36.

\bibitem{Zhang D}
Zhang D, Chen L, Zhang S, Xu H, Zhao Z, Yu K. 
\newblock Large Language Models Are Semi-Parametric Reinforcement Learning Agents. 
\newblock Advances in Neural Information Processing Systems. 2024 Feb 13;36.

\bibitem{Carta T}
Carta T, Romac C, Wolf T, Lamprier S, Sigaud O, Oudeyer PY. 
\newblock Grounding large language models in interactive environments with on-policy reinforcement learning. 
c In International Conference on Machine Learning 2023 Jul 3 (pp. 3676-3713). PMLR.

\bibitem{Yang S}
Yang S, Gribovskaya E, Kassner N, Geva M, Riedel S. Do Large Language Models Latently Perform Multi-Hop Reasoning?. arXiv preprint arXiv:2402.16837. 2024 Feb 26.

\bibitem{Zhao Z}
Zhao Z, Lee WS, Hsu D. 
\newblock Large language models as commonsense knowledge for large-scale task planning. 
\newblock Advances in Neural Information Processing Systems. 2024 Feb 13;36.

\bibitem{Brockman}
Brockman G, Cheung V, Pettersson L, Schneider J, Schulman J, Tang J, Zaremba W. Openai gym. arXiv preprint arXiv:1606.01540. 2016 Jun 5.

\bibitem{Florensa}
Florensa C, Held D, Wulfmeier M, Zhang M, Abbeel P. Reverse curriculum generation for reinforcement learning. InConference on robot learning 2017 Oct 18 (pp. 482-495). PMLR.

\end{thebibliography}

\end{document}